\documentclass{article}

\usepackage{microtype}
\usepackage{graphicx}
\usepackage{subfigure}
\usepackage{booktabs} 

\usepackage{hyperref}

\usepackage[accepted]{icml2023}

\usepackage{amsmath}
\usepackage{amssymb}
\usepackage{mathtools}
\usepackage{amsthm}

\usepackage[capitalize,noabbrev]{cleveref}

\theoremstyle{plain}

\theoremstyle{definition}

\theoremstyle{remark}

\usepackage[textsize=tiny]{todonotes}

\icmltitlerunning{Action and Trajectory Planning using Hierarchical Reinforcement Learning}

\begin{document}

\twocolumn[
\icmltitle{Action and Trajectory Planning for Urban Autonomous Driving with Hierarchical Reinforcement Learning}



\icmlsetsymbol{equal}{*}

\begin{icmlauthorlist}
\icmlauthor{Xinyang Lu}{nus}
\icmlauthor{Flint Xiaofeng Fan}{nus,astar}
\icmlauthor{Tianying Wang}{astar}
\end{icmlauthorlist}

\icmlaffiliation{nus}{National University of Singapore, Singapore}
\icmlaffiliation{astar}{Institute of High-Performance Computing, A*STAR, Singapore}

\icmlcorrespondingauthor{Xinyang Lu}{xinyang.lu@u.nus.edu}

\icmlkeywords{Machine Learning, ICML}

\vskip 0.3in
]



\printAffiliationsAndNotice{\icmlEqualContribution} 

\begin{abstract}
Reinforcement Learning (RL) has made promising progress in planning and decision-making for Autonomous Vehicles (AVs) in simple driving scenarios.  
However, existing RL algorithms for AVs fail to learn critical driving skills in complex urban scenarios. 
First, urban driving scenarios require AVs to handle multiple driving tasks of which conventional RL algorithms are incapable.
Second, the presence of other vehicles in urban scenarios results in a dynamically changing environment, which challenges RL algorithms to plan the action and trajectory of the AV.
In this work, we propose an \textbf{a}ction and \textbf{t}rajectory planner using \textbf{H}ierarchical \textbf{R}einforcement \textbf{L}earning (\textbf{atHRL}) method, which models the agent behavior in a hierarchical model by using the perception of the lidar and birdeye view. 
The proposed atHRL method learns to make decisions about the agent's future trajectory and computes target waypoints under continuous settings based on a hierarchical DDPG algorithm. The waypoints planned by the atHRL model are then sent to a low-level controller to generate the steering and throttle commands required for the vehicle maneuver.
We empirically verify the efficacy of atHRL through extensive experiments in complex urban driving scenarios that compose multiple tasks with the presence of other vehicles in the CARLA simulator. The experimental results suggest a significant performance improvement compared to the state-of-the-art RL methods.
\end{abstract}

\section{Introduction}

Autonomous driving has been greatly advanced with the rapid development of machine learning in the past few years. The autonomous driving system can be roughly categorized into several different parts: perception, localization, planning, and control. The planning module is one of the most challenging tasks in achieving a reliable autonomous driving system, especially when experiencing dense traffic with a large number of diverse participants. 

Traditional planners for autonomous driving could be roughly categorized into optimization-based planners and sampling-based planners~\citep{paden2016a}. They are usually designed based on rules and strongly rely on the prediction and sensing information, thus they are unable to generalize or handle interactive scenarios well. To address the uncertainties and changes in the intentions during the interaction process in those scenarios, many learning-based planners have been proposed in recent years \citep{zhu2021a,kiran2021deep}.

More recently, Hierarchical Reinforcement Learning (HRL) based planners have been proposed for autonomous driving. Compared to directly outputting control policies such as steering and throttle, HRL-based methods better model the multi-layer decision-making process in driving, where the lower-level decisions are dependent on high-level decisions. In \cite{naveed2020trajectory}, an option of either lane change or lane follow is provided by the high-level planner, and the mid-level trajectory planner then learns to output waypoints for the vehicle to follow. As a recent analysis in HRL autonomous driving, \cite{qiao2020behavior} proposes a three-layer HRL method based on Deep Q Network (DQN) \cite{mnih2015human} and handcraft some decision choices to plan the behavior and trajectory in urban driving scenarios, achieving promising results in the lane changing and left-turn scenarios. However, the discrete action space in Q-learning-based methods imposes additional constraints on the choices of available driving decisions in complex urban driving scenarios. With experiments, we show that current H-DQN methods like \cite{qiao2020behavior} are struggling with driving scenarios involving multiple tasks and with the presence of other vehicles.
\begin{figure}[t]
    \centering
    \includegraphics[width=0.45\textwidth]{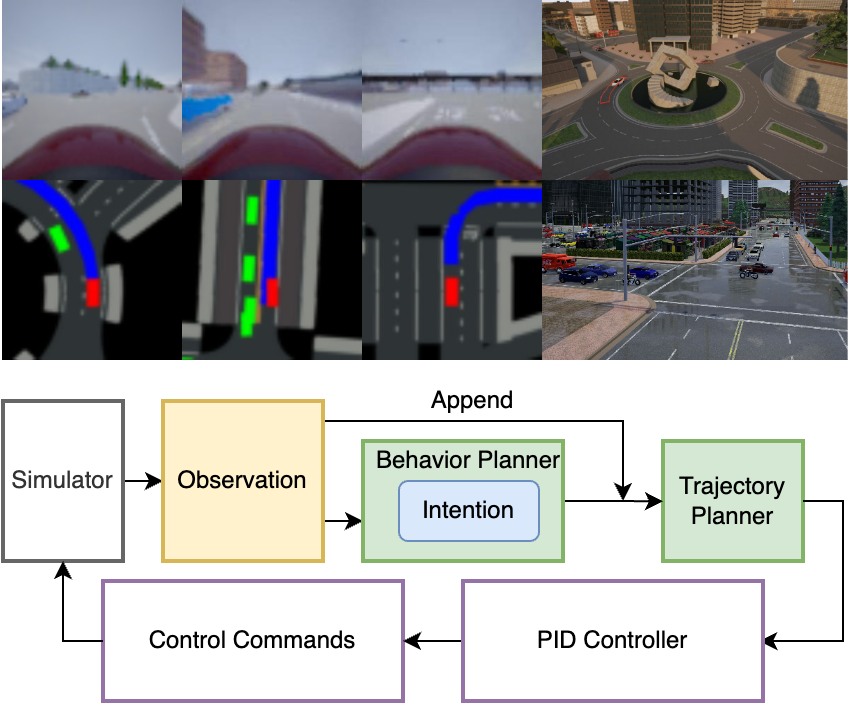}
    \caption{The architecture of the proposed hierarchical system}
    \label{fig:architecture}
\end{figure}

In this work, we propose an \textbf{a}ction and \textbf{t}rajectory planner using \textbf{H}ierarchical \textbf{R}einforcement \textbf{L}earning (\textbf{atHRL}) to model the hierarchical behaviors of decision making and planning for autonomous driving, as shown in Fig.~\ref{fig:architecture}. In the proposed method, a hierarchical architecture is adopted which includes a high-level intention planner learning a driving intention, a mid-level trajectory planner combining the intention with the original observation to learn the trajectory and the desired speed, and a low-level PID controller to compute the actions of throttle and steering of the vehicle in the environment from the trajectory and speed. The two RL planners both adopt the off-policy actor-critic Deep Deterministic Policy Gradient (DDPG) \cite{lillicrap2019continuous} algorithm. The PID controller is then used to compute the motion command to track the waypoints. The proposed atHRL explicitly models a three-layer hierarchical decision and planning process with dedicated information flow and thus achieves better and more reliable performance. Moreover, hierarchical DDPG methods demonstrate superior efficacy in larger and more complex environments, as they possess the capability to make decisions in a continuous space. In contrast, Q-learning-based methods are limited by their discrete decision-making approach. With experiment results conducted in dynamic urban driving scenarios in the CARLA simulator that consist of multiple driving tasks and involve other vehicles, we empirically verify that our method outperforms conventional RL planners and other similar HRL-based planners such as method based on hierarchical DQN in \cite{qiao2020behavior}, which indicates that the hierarchical off-policy actor-critic planner suits the decision-making tasks in urban autonomous driving better.

The main contributions are summarized below:
\begin{itemize}
    \item we propose a three-level hierarchical structure to model the planning and decision-making process in urban autonomous driving environments with the mid-level perception data.
    \item we use two off-policy actor-critic structures to learn two decision layers that generate trajectory in continuous space and apply a PID controller to compute the throttle and steering. 
    \item we propose the \textbf{atHRL} algorithm that makes better performance compared to alternative  RL and Q-learning-based HRL methods in dynamic urban autonomous driving scenarios due to the robustness and smoothness of driving in the continuous action space
\end{itemize}

\begin{figure*}[t]
    \centering
    \includegraphics[width=0.9\textwidth]{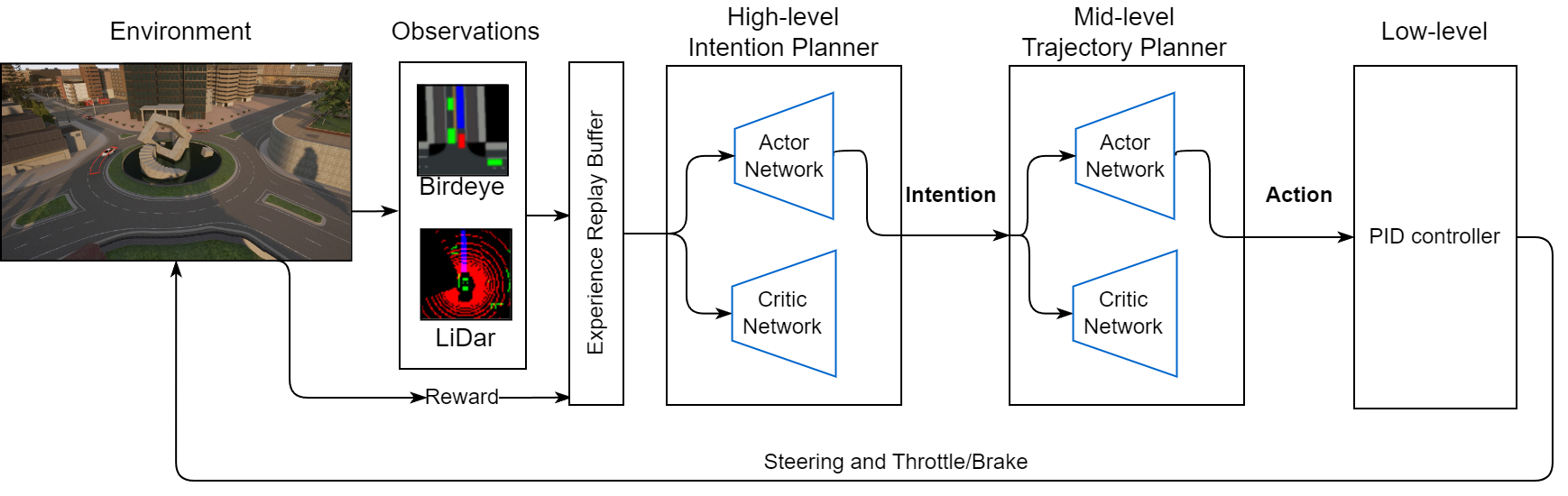}
    \caption{The overview of the proposed atHRL planner: two DDPG agents are used to learn the intention and a continuous trajectory, and a PID controller is used to compute the steering and throttle based on the waypoints.}
    \label{fig:workflow}
\end{figure*}
\section{Related Work}

\subsection{Reinforcement Learning}

Reinforcement Learning has been proven to be a promising technique in training agents to handle various tasks through learning from observations. In the past few years, a number of RL algorithms have been proposed, including value-based methods such as DQN \cite{mnih2015human} and its variants, policy-based methods such as A3C \cite{mnih2016asynchronous}, DDPG \cite{lillicrap2019continuous} and PPO \cite{schulman2017proximal}. Traditional RL methods have been successfully applied to complex decision-making tasks in different fields and have accomplished a lot. For example, early in 2015, \cite{mnih2015human} applied DQN to train agents to play Atari 2600 games and achieved good performance over a set of games. More recently, \cite{vinyals2019grandmaster} successfully trained an agent to master StarCraft with reinforcement learning. In robotics, \cite{levine2016end} and \cite{gu2016deep} developed  reinforcement learning methods and learned policies to control the robotic arms to perform different tasks. In addition, RL-based methods have also been used for autonomous driving tasks and demonstrated several successes in several scenarios~\cite{zhu2021a,kiran2021deep,8793742}. For example, \cite{8461233} adopted DQN to learn policies to deal with the intersection scenario and outperforms heuristic approaches. However, many RL methods suffer from sample inefficiency in practical applications \cite{fan2021fault,fan2023fedhql} and are only capable of handling simple tasks such as lane changing or protected turning and might perform badly in dynamic and complex scenarios comprised of several different tasks.

\subsubsection{Q Learning}

As one of the most famous RL algorithms, Q learning \cite{qlearning} is a value-based method that aims to optimize the action-value function \(Q^*(s, a)\) typically by updating the Bellman Equations:

\begin{equation}
V(s)=E[r(s, a)+\gamma V(s')]
\end{equation}

\begin{equation}
Q(s, a)=E[r(s, a)+\gamma E[Q(s', a')]],
\end{equation}

where $s, a$ are the state and action, $s', a'$ are the next state and the corresponding action. The updates of Q learning are in an off-policy manner, which means that at each time step any data collected in the past can be used for learning. Over the years, Q learning has been proven to be a very useful reinforcement learning algorithm, but it can generally only deal with discrete action spaces and thus has poor performance in high-dimension spaces, especially in continuous and dynamic environments. 

\subsubsection{Policy Gradient} 

In contrast to Q learning, policy gradient (PG) methods are policy-based. They use gradient descent to optimize the parameterized policy \(\pi_\theta\). With \(J(\pi_\theta)\) representing the return of the policy, the gradient of it with respect to \(\theta\) can be calculated by:
\begin{equation}
\nabla_\theta J(\pi_\theta) = \underset{\tau \sim \pi_\theta}{E}[\sum_{t=0}^{T} \nabla_\theta log\pi_\theta(a_t|s_t)A^{\pi_\theta}(s_t|a_t)],
\end{equation}

where $s_t$ and $a_t$ represents the state at $t$ and the action at state $t$, $\tau$ is the trajectory and $A$ is the advantage function.
Thereafter, the policy parameters can be updated using stochastic gradient descent:
\begin{equation}
\theta_{k+1} = \theta_k + \alpha \nabla_\theta J(\pi_\theta)
\end{equation}

Compared to Q learning methods, PG methods directly optimize the policy return and thus converge to a good policy in a more stable manner. In addition, high dimensional and continuous action spaces can be dealt with in PG methods, which Q learning methods generally cannot handle. Specifically, in complex and dynamic environments as in our experiment, traditional Q learning methods may perform not as well as PG methods.

\subsubsection{Hierarchical Reinforcement Learning} 

In recent years, advances in Hierarchical Reinforcement Learning (HRL) have greatly improved the ability of RL agents to make more complicated decisions and achieve difficult tasks more efficiently. HRL decomposes RL problems into higher-level and lower-level sub-tasks which are solved separately using RL algorithms. This approach reduces the computational complexity by solving several less difficult sub-tasks and hence can handle more difficult tasks than traditional RL methods. \cite{8793742} proposes h-DQN that generates a sub-goal by the meta-controller and is used to guide the low-level actions. Later \cite{nachum2018data} extends h-DQN to an off-policy framework with the HIRO algorithm. With hindsight action and transitions, HAC \cite{levy2019learning} is able to learn multiple levels of policies in parallel and accelerate the learning process. Moreover, \cite{nair2018overcoming} applied the DDPG algorithm with Hindsight Experience Replay to solve complicated robotic tasks in continuous action space. HRL has fully demonstrated the ability to handle complex tasks with a multi-layer decision model. Recently, a few works have been attempting to apply HRL to autonomous driving applications to model the hierarchical decision-making structure, such as ~\cite{qiao2020behavior,qiao2020hierarchical} with hierarchical DQN and \cite{Duan_2020} with asynchronous parallel HRL, but only apply to a few specific tasks and may not adapt to a different type of tasks.

\subsection{Planning and Decision Making in Autonomous Driving}

Many previous works have been conducted for planning and decision-making in autonomous driving applications.
Several existing methods model the decision-making tasks as Partially Observable Markov Decision Process (POMDP). For example, \cite{6957722} presents a continuous space POMDP model that controls the velocity of the vehicle in a merging scenario. A multi-policy decision-making algorithm~\cite{7139412} evaluates and selects the best policy under different situations to handle uncertain and dynamic environments. After that, \cite{8911507} combines RL and the planning methods from POMDP, using a Monte Carlo tree search to select appropriate actions while training the RL agent, which performs well in various highway scenarios.

Imitation learning that intends to mimic policies from expert experience proves to be an effective way to train agents in a structured environment. \cite{codevilla2018endtoend} applies conditional imitation learning while allowing manual control of the vehicle. ChauffeurNet \cite{bansal2018chauffeurnet} further improves the simple imitation by designing additional handcrafted losses and adding synthesized perturbations to handle the distribution drift. However, it is still highly dependent on the amount and the quality of the provided expert data, and not able to fully explore the varieties of intentions during the driving. 

In addition, agents trained with RL are earning more attention in the field. With a well-designed architecture, deep RL agents are able to summarize a viable policy rapidly by learning from successful and unsuccessful trials. \cite{6856582} combines several planners to achieve smooth maneuvers and uses a PCB algorithm to coordinate the throttle and steering controllers of the vehicle. A continuous decision-making module with a three-stage policy that interacts with the dynamic environment is implemented in \cite{8500605}. The module shows its potential in stabilizing learning when dealing with complicated driving tasks. Furthermore, trials with constructing HRL structures are emerging as a trend in recent analyses. \cite{8500368} proposed to learn the hierarchical policies with HRL to deal with decision-making in autonomous driving scenarios, which proves to be more effective than non-hierarchical planners. \cite{9241055} adopted the hierarchical idea to design the agent for autonomous driving with a high-level decision-making model and a low-level motion planning model. It is shown that good results have been achieved in many specific scenarios such as lane changing and turning. However, these analyses have not applied to a complete urban driving environment. In our work, the HRL agent will be operating in a complete map consisting of various tasks with the presence of other vehicles, which means that different policies need to be learned at the same time.


\section{METHODOLOGY}

\subsection{Problem Formulation} 

\subsubsection{Preliminary on MDP and DDPG}

The autonomous driving tasks discussed in this paper are formulated as Markov Decision Processes (MDPs) $M = <S, A, T, R, \gamma> $, where $S$ is a set of states $s$, $A_i$ is a set of actions $a$; $T_s$ is the state transition; $R: S \times A \longrightarrow R$ is the set of rewards $s$; $\gamma \in [0,1]$ is the discount factor. The total episodic reward is then the summation of discounted rewards: $r_{total} = \sum_{i=0}^{T_e} \gamma^i r_i$, where $T_e$ is the total steps of an episode.

Recall that, the goal in RL is to find the optimal policy that maximizes the expected reward. This optimal action-value function can be described by the Bellman equation:

\begin{equation}
Q^* \left( s, a\right) = \mathop{E}\limits_{s'\sim P}[r\left(s, a\right)+\gamma \mathop{\max}\limits_{a'} Q^*\left(s', a'\right)],
\end{equation}

where $s'\sim P$ means that the next state $s'$ is sampled from a distribution $P(\cdot|s, a)$ from the environment. As discussed in the previous section, policy gradient methods perform better than traditional Q-learning methods in complex and dynamic environments in our experiment. We consider the off-policy actor-critic RL algorithm Deep Deterministic Policy Gradient (DDPG) for this task, which is based on Q-learning but solves continuous action space problems. DDPG aims to minimize the mean squared Bellman error (MSBE) which represents the error between the Q-value and the Bellman equation: 

\begin{equation}
\resizebox{.9\hsize}{!}{$L\left( \phi, D \right)=\mathop{E}\limits_{\left( s, a, r, s', d \right)\sim D}\left[ \left( Q_\phi\left( s, a\right)-\left( r+\gamma \left( 1-d\right) \mathop{\max}\limits_{a'} Q_\phi\left( s', a'\right) \right) \right)^2\right],$}
\end{equation}

where $D$ is the collection of the transitions $(s,a,r,s',d)$ and $d$ indicates the terminal state. DDPG deploys the main network with actor and critic networks while maintaining a target network to stabilize the Q-learning process in a continuous domain, where the target is the term in the MSBE loss:

\begin{equation}
\mathcal{L} = r+\gamma \left( 1-\!d\right) \mathop{\max}\limits_{a'} Q_\phi\left( s', a'\right) 
\end{equation}

The target network is updated at a slower frequency to maintain the training stability:

\begin{equation}
\phi_{targ} \leftarrow \rho \phi_{targ} + \left( 1 - \rho \right) \phi,
\end{equation}
where $\phi$ is a hyperparameter with a value between 0 and 1.

\subsubsection{Formulation of the HRL} 

We further design the task observations, actions, and rewards to formulate the HRL problem by modeling the planning and decision process in a multi-level structure.

\textit{Observations:} we use mid-level perception including lidar perceptions and BEV images as the observations, similar to \cite{chen2019model}. 

\textit{Actions for high-level intention:} the high-level action is a latent representation of intentions. It is a value $I$ in the range from $0$ to $2$, which represents the options of going left, going straight, and going right that are to be passed to the trajectory planner to decide the range of the waypoints.

\textit{Actions for mid-level trajectory:} the mid-level action is a trajectory represented by both waypoints and desired speed at the point. In this work, since we use a PID controller to track the waypoints, we only output one waypoint and the corresponding desired speed for simplicity. 

\textit{Rewards:} the design of the reward is based on a few factors, including longitudinal speed, penalties that come from collisions, out-of-lane, large steering angle, exceeding the speed limit, and large lateral acceleration. It is designed based on the one used in \cite{chen2019model} which used a similar environment setting. The overall reward function is as follows:

\begin{align}
    r = \alpha_1 r_{collision} + \alpha_2 r_{longspeed} +  \alpha_3 r_{exceed} \nonumber \\
    + \alpha_4 r_{out} + \alpha_5 r_{steer} + \alpha_6 r_{latspeed} + \alpha_7
\end{align}

In our work, the corresponding weights of each factor are designed as follows: $\alpha_1=200, \alpha_2=1, \alpha_3=10, \alpha_4=1, \alpha_5=5, \alpha_6=0.2, \alpha_7=-0.1$.

\subsection{atHRL: Hierarchical Driving Model for Planning}

\begin{figure}[t]
    \centering
    \includegraphics[width=0.4\textwidth]{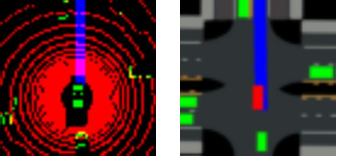}
    \caption{A pair of lidar and birdeye images, which are used as the input observations of the neural network}
    \label{fig:observation}
\end{figure}

\subsubsection{Overview of atHRL}

As shown in Fig.~\ref{fig:workflow}, we propose a three-level hierarchical structure with a high-level intention planner, a mid-level trajectory planner, and a low-level PID controller to generate the motion command for trajectory. The intention planner and the trajectory planner are two DDPG agents with Long Short Term Memory (LSTM) arranged in a hierarchical reinforcement learning architecture. 

\subsubsection{The Neural Network}

2D lidar image and BEV image are used as observation inputs as shown in Fig.~\ref{fig:observation}. The dimension of the 2D lidar image is $ 32 \times 32 \times 1$, which represents point clouds projected on the map; and the dimension of the BEV semantic image is $ 32\times 32 \times 3$, showing the map, the surrounding objects, and the route of ego vehicle. With the given observations, the planner computes an intention from the set of ranges. The intention serves as an intermediate value in the neural network and will be concatenated with the observation and passed to the trajectory planner. It gives extra information on top of the original sensor observations, and the trajectory planner is only soft-constrained and guided by the intention. 

\begin{figure}[t]
    \centering
    \includegraphics[width=0.4\textwidth]{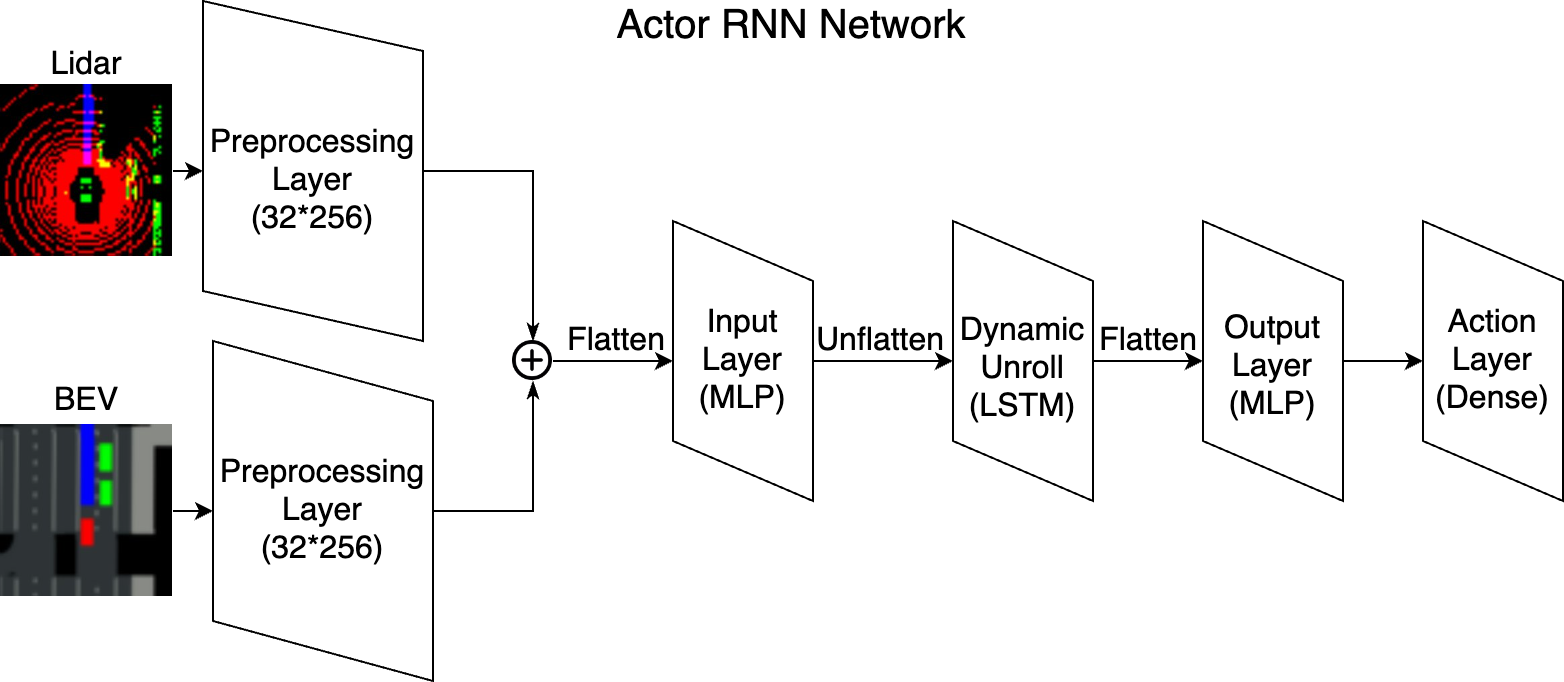}
    \caption{The neural network for computing the intention and trajectory}
    \label{fig:network}
\end{figure}

The general neural network structure in the intention and trajectory planners is as shown in Fig.~\ref{fig:network}. It takes the 2D lidar images and BEV semantic images as observations which go through the $32 \times 256$ preprocessing layers and are concatenated together. Then, the network consists of an Input Layer of Multilayer Perceptron (MLP), an LSTM layer, an Output Layer of MLP, and a Dense Layer for the output of actions.

Afterward, the trajectory planner computes the desired speed and the trajectory of target waypoints for the vehicle to drive to. The desired speed is based on the speed limit defined in the simulator and the target waypoint is restricted to a fixed-sized semi-circle in front of the vehicle, assuming that the vehicle would not drive reversely in the scenario. This is to simulate the decision-making of human drivers, where a place to go and how fast to drive is decided before drivers control the throttle and steering. Therefore, the algorithm could well handle dynamic tasks in different complex urban driving scenarios. 

In the end, the PID controller generates the steering angle given the target waypoint predicted by the trajectory planner under the local coordinate of the ego vehicle and calculates the control of the throttle brake based on the desired speed. The steering angle and the control of the throttle and brake then serve as the direct command provided to the agent's vehicle in the environment. 

\subsubsection{Hierarchical Off-Policy Actor-Critic}

\begin{algorithm}[tb]
    \caption{atHRL Planner Algorithm}
    \label{alg:planner}
    \begin{algorithmic}[1]
        \STATE Initialize action planner and trajectory planner $P^i$ and $P^t$ with actor RNN networks $R^{ai}$, $R^{at}$ and critic RNN networks $R^{ci}$, $R^{ct}$.
        \STATE Initialize the replay buffer $B$.
        \STATE Run N steps with random policy to collect experience and store in buffer $B$.
        \FOR{N+1 to S steps} 
            \STATE Get initial state $s_0$ from environment
            \WHILE{$s_0$ is not terminal state}
            \STATE Select intention $i$ with the intention planner $P^i$ where observation $O = \{birdeye, lidar\}$
            \STATE Select target speed $v$ and waypoint $w$ with the trajectory planner $P^t$ where observation $O =\{birdeye, lidar, i\}$
            \STATE Calculate \textit{Throttle} $= PID_{longi}(u, v)$, where u is the current speed 
            \STATE Calculate \textit{Steering = $PID_{late}(w)$}, using local coordinate
            \STATE Step in the environment and get the next state $s_{t+1}$ and reward $r$
            \STATE Store transitions into buffer $B$
            \ENDWHILE
            \STATE Train DDPG agents with buffer $B$, and update the network
        \ENDFOR
    \end{algorithmic}
\end{algorithm}

The training of the off-policy actor-critic RL algorithm is similar to \cite{chen2019attention,qiao2020hierarchical}. The complete method flow to train the atHRL system is described in Algorithm \ref{alg:planner}. The mid-level DDPG agent computes the mid-level rewards $r_l$, and computes the TD error. The high-level DDPG agent computes the high-level rewards $r_h$ while back-propagating both high-level and mid-level TD errors. Since this is an off-policy RL algorithm, a reply buffer is used to store the interaction for off-policy training.


\section{EXPERIMENT AND RESULTS}

\subsection{Environment Setup}

\begin{figure}[t]
    \centering
    \includegraphics[width=0.45\textwidth]{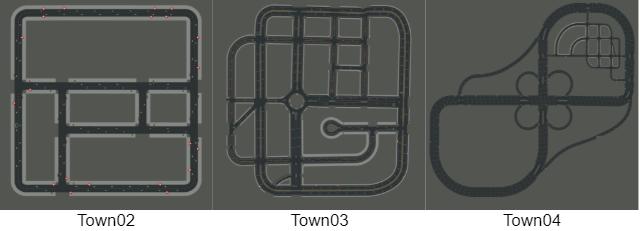}
    \caption{The selected test maps from the CARLA simulator}
    \label{fig:maps}
\end{figure}

The experiments are conducted using CARLA simulator~\cite{dosovitskiy2017carla} with OpenAI Gym interface, following the setting in \cite{chen2019model}. The simulator provides an urban driving environment to train the RL agents effectively and allows users to add other vehicles that can autonomously interact with the agents. To fully investigate the ability of the vehicle to drive in various urban environments, we selected three maps from the simulator. Each map is composed of multiple tasks to test the ability of the vehicle to learn to plan under various situations, as shown in Fig.~\ref{fig:maps}. For example, the map Town02 is mainly made up of T-turnings and does not consist of many complex situations. Meanwhile, there are a lot of long straight roads in the map Town04, as well as a smaller town part at the top right corner with multiple turnings inside. In contrast, the map Town03 is more complex, consisting of several different driving situations such as T-turnings, long straight roads, a roundabout, and a five-lane junction. 
The experiments are conducted on each map separately. The diversity of the tasks in each map can help to verify whether the RL agent can adapt to handle different situations. 

For each scenario, we also add 100 background traffic vehicles, which is the same amount as in the experiments in \cite{chen2019model}. The vehicles will automatically move around the map and will interact with the RL agent's vehicle, randomly resulting in different driving tasks with interaction with other vehicles. The RL-controlled ego-vehicle thus needs to learn to avoid collisions, stay in lane and interact with other vehicles properly. This creates a dynamic environment and increases the difficulty for the RL agents to learn. For each map, we start the experiment by initializing the experience replay buffer by running a random policy for 10,000 steps, followed by a training of 30,000 steps. In addition, we compare the proposed method with several baseline algorithms to validate its effectiveness.

\begin{figure}[t]
    \centering
    \includegraphics[width=0.45\textwidth]{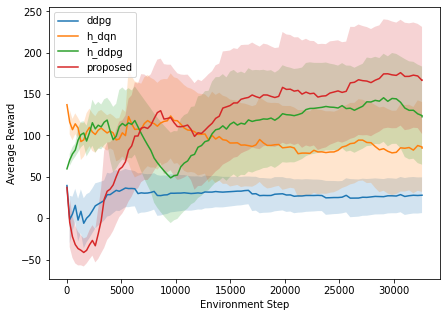}
    \caption{Training reward of the different methods.}
    \label{fig:reward}
\end{figure}

\begin{figure}[t]
    \centering
    \includegraphics[width=0.23\textwidth]{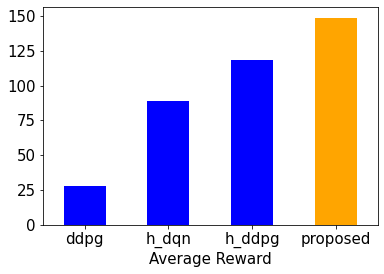}
    \includegraphics[width=0.22\textwidth]{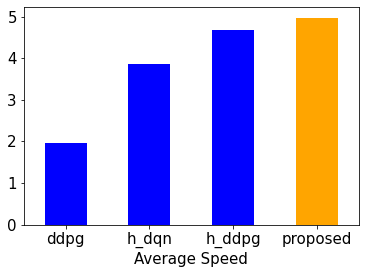}
    \caption{Results comparison in terms of average reward and average speed in map Town03.}
    \label{fig:bar}
\end{figure}
  
\subsection{Results and Discussion}

In the experiments, we compare the proposed method (atHRL) with three other HRL and RL methods, including hierarchical DQN, original DDPG, and a two-layer hierarchical DDPG. The hierarchical DQN method proposed in \cite{qiao2020behavior} is an HRL algorithm that also adopted trajectory planning and is based on DQN with discrete action output. It performs well in simpler driving scenarios with few decision-making options. We would like to use it as a comparison between our proposed method and the conventional HRL methods with discrete action spaces. The original non-hierarchical DDPG method and a two-layer hierarchical DDPG that does not include the intention planner are taken as examples to validate the effectiveness of our three-layer hierarchical DDPG method. 

Fig.~\ref{fig:reward} indicates the reward of the different methods during the training steps. The reward takes into account the penalties of collisions, out-of-lane, large steering angle exceeding the speed limit, and large lateral acceleration, hence the comparisons of the reward represent well the performance comparisons. It shows that after the algorithms converge after the 30,000 training steps, our proposed method has the highest reward among the four methods. Fig.~\ref{fig:bar} shows that our proposed method outperforms all other methods in the experiment in terms of both total average reward and total average speed. While the average reward indicates the overall performance of the agents, the average speed can help to verify the soundness of the vehicles' driving capabilities.

\begin{figure}[t]
    \centering
    \includegraphics[width=0.23\textwidth]{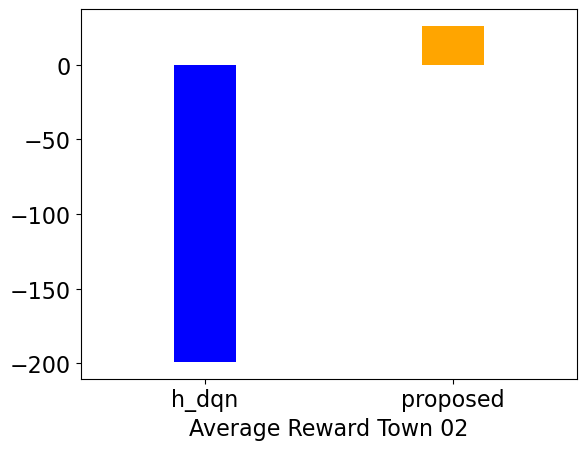}
    \includegraphics[width=0.23\textwidth]{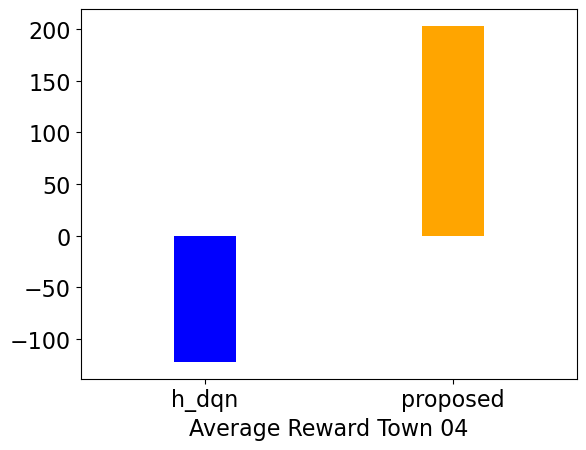}
    \caption{The performance of the original DDPG method and our proposed method in other maps. The left graph is on map Town02 and the right graph is on map Town04}
    \label{fig:bar_t2_t4}
\end{figure}

The direct comparisons of our proposed method with the traditional non-hierarchical DDPG algorithm in the map Town03 show that the adoption of hierarchical planners improves the overall performance of the reinforcement learning agents in the selected urban driving scenarios. This is not only reflected by the higher total average reward, which indicates the agent's ability to guarantee safety by avoiding collisions and staying in the simulator for a long time but also by the higher average driving speed which indicates its robustness of handling complex situations. This shows that the high-level decision-making planners enhance the stability of the motion control and therefore avoid collisions. Meanwhile, learning low-level control commands directly from observations may cause instability of controls and difficulty in learning different policies of different tasks in the same scenario, hence should be avoided. Furthermore, the comparison between our proposed method and the two-layer trajectory planning H-DDPG shows that the higher level intention planner is able to help the agent to make smarter decisions on the choice of trajectory and hence can achieve better performances. In addition, our method outperforms the hierarchical DQN with a similar trajectory planning structure proposed in \cite{qiao2020behavior}. While H-DQN with discrete action spaces is capable of handling simpler decision-making choices, it performs badly in more complex urban driving environments with combinations of different tasks. The actor-critic planner in our method, however, allows continuous decision-making and improves the overall performance in policy learning, hence leading to better results in complex urban driving scenarios.

As shown in Fig.~\ref{fig:bar_t2_t4}, the average reward of our proposed method is higher than the hierarchial DQN method in the other maps, Town02 and Town04 as well. Considering that the driving scenarios in the three maps are significantly different from each other, the generally better performance of our method indicates that this algorithm is capable of adapting to various situations and achieving more robust performances compared to the baseline HRL methods in different environments. This significant improvement is due to the improvement from limited and discrete selections of waypoints to continuous ones for the low-level controllers to follow. In multi-task and dynamic urban driving scenarios, the performance of Q-learning-based methods is limited by the discrete action space. 

\begin{figure}[t]
    \centering
    \includegraphics[width=0.45\textwidth]{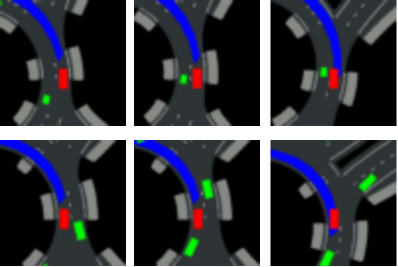}
    \caption{Example cases of meeting other vehicles when passing the roundabout. The upper cases are taken from the original DDPG and the lower cases are from the proposed atHRL methods.}
    \label{fig:roundabout}
\end{figure}

\subsection{Case Studies}

To analyze how the vehicles in our method operate, a few representative scenarios are discussed. For example, Fig.~\ref{fig:roundabout} shows a complicated roundabout scenario where the agent needs to avoid collision with other vehicles when passing through the roundabout. This case compares the behaviors of the agent trained by the original DDPG method with the agent from our proposed method. The agent from conventional H-DQN methods could not make meaningful movements under this scenario at all, and thus is not included in the comparison. From the upper three figures, it can be seen that the DDPG agent tries to make way for the coming vehicle behind it by slowing down its speed, but its steering angle still leads to the collision with the coming vehicle. In contrast, the agent trained with our proposed method maintains smooth maneuvering while keeping a reasonable speed. The high-level decision-making layers have planned a well-selected continuous trajectory for the low-level controllers to follow, thus the tracking is easier and more stable. On the other hand, the H-DQN agent and the DDPG agent either do not have a pre-determined continuous trajectory to follow or the driving can be easily influenced by subtle changes in the determination of steering angles, hence failing to perform well in dynamically changing scenarios that involve interaction with other vehicles. 

\begin{figure}[t]
    \centering
    \includegraphics[width=0.5\textwidth]{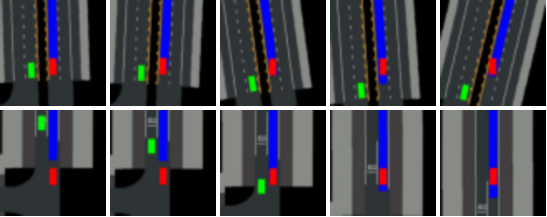}
    \caption{Case studies of going straight when meeting other vehicles. The upper cases are taken from the original DDPG and the lower cases are from the proposed atHRL methods.}
    \label{fig:straight}
\end{figure}

Fig.~\ref{fig:straight} illustrates another example where the agent performs a simple task of going straight while meeting vehicles from another side of the road and no lane change is needed to be performed in this case. As is shown in the upper figures, the DDPG agent is unstable and can hardly maintain a straightforward driving direction. In contrast, the agent of our proposed method is able to drive in a fixed direction. The high-level intention and trajectory planners have stabilized the movements of the agent and made the driving in dynamic scenarios more robust. Besides, the PID controller that follows the pre-determined trajectory guarantees smoothness and safety in the movements compared to the direct generation of steering angles in the DDPG agent.

\begin{figure}[t]
    \centering
    \includegraphics[width=0.5\textwidth]{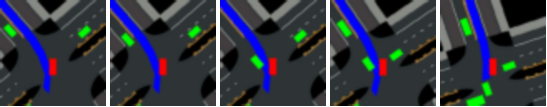}
    \caption{Case studies of unprotected left-turn at the intersection.}
    \label{fig:turn}
\end{figure}

Furthermore, Fig.~\ref{fig:turn} shows another case where the agent of our proposed method performs an unprotected left turn at an intersection while many other vehicles are present at the intersection as well. The completion of this task further proves that our proposed method is able to deal with very complex and dynamic driving scenarios robustly and smoothly where the movements of other vehicles are unpredictable. When adopting conventional RL methods, it is often observed that their agents cannot maintain a smooth movement in the turning and are likely to collide with other vehicles.
To summarize, these case studies validate the more stable control from our hierarchical architecture compared to non-hierarchical methods.


\section{CONCLUSION}

In this work, we propose atHRL, an action and trajectory planner using a Hierarchical Reinforcement Learning algorithm for complex driving tasks in multiple dynamic urban scenarios. The proposed method adopts the DDPG algorithm with a hierarchical structure to learn the action and trajectory, which better models the human decision process and achieves robust and smooth control of the vehicle in the continuous action space. The experimental results indicate that while other RL methods including conventional DDPG, two-level Hierarchical DDPG, and three-level Hierarchical DQN \cite{qiao2020behavior} fail to perform well in complex urban driving scenarios that contain multiple driving tasks and interactions with other vehicles, our proposed method is able to perform reliable driving behaviors and achieve better results. 

Meanwhile, there have been many other interesting policy optimization reinforcement learning algorithms. In the current work, we have only applied our hierarchical architecture with the DDPG algorithm. The extension of the architecture to other reinforcement learning algorithms can be an interesting research direction that is worth exploring in our future work.


\bibliography{example_paper}

\begin{thebibliography}{34}
\providecommand{\natexlab}[1]{#1}
\providecommand{\url}[1]{\texttt{#1}}
\expandafter\ifx\csname urlstyle\endcsname\relax
  \providecommand{\doi}[1]{doi: #1}\else
  \providecommand{\doi}{doi: \begingroup \urlstyle{rm}\Url}\fi

\bibitem[Bansal et~al.(2018)Bansal, Krizhevsky, and
  Ogale]{bansal2018chauffeurnet}
Bansal, M., Krizhevsky, A., and Ogale, A.
\newblock Chauffeurnet: Learning to drive by imitating the best and
  synthesizing the worst, 2018.

\bibitem[Brechtel et~al.(2014)Brechtel, Gindele, and Dillmann]{6957722}
Brechtel, S., Gindele, T., and Dillmann, R.
\newblock Probabilistic decision-making under uncertainty for autonomous
  driving using continuous pomdps.
\newblock In \emph{17th International IEEE Conference on Intelligent
  Transportation Systems (ITSC)}, pp.\  392--399, 2014.

\bibitem[Chen et~al.(2018{\natexlab{a}})Chen, Tang, Xin, Li, and
  Tomizuka]{8500605}
Chen, J., Tang, C., Xin, L., Li, S.~E., and Tomizuka, M.
\newblock Continuous decision making for on-road autonomous driving under
  uncertain and interactive environments.
\newblock In \emph{2018 IEEE Intelligent Vehicles Symposium (IV)}, pp.\
  1651--1658, 2018{\natexlab{a}}.

\bibitem[Chen et~al.(2018{\natexlab{b}})Chen, Wang, and Tomizuka]{8500368}
Chen, J., Wang, Z., and Tomizuka, M.
\newblock Deep hierarchical reinforcement learning for autonomous driving with
  distinct behaviors.
\newblock In \emph{2018 IEEE Intelligent Vehicles Symposium (IV)}, pp.\
  1239--1244, 2018{\natexlab{b}}.

\bibitem[{Chen} et~al.(2019){Chen}, {Yuan}, and {Tomizuka}]{chen2019model}
{Chen}, J., {Yuan}, B., and {Tomizuka}, M.
\newblock Model-free deep reinforcement learning for urban autonomous driving.
\newblock In \emph{2019 IEEE Intelligent Transportation Systems Conference
  (ITSC)}, pp.\  2765--2771, 2019.

\bibitem[Chen et~al.(2019)Chen, Dong, Palanisamy, Mudalige, Muelling, and
  Dolan]{chen2019attention}
Chen, Y., Dong, C., Palanisamy, P., Mudalige, P., Muelling, K., and Dolan,
  J.~M.
\newblock Attention-based hierarchical deep reinforcement learning for lane
  change behaviors in autonomous driving.
\newblock In \emph{Proceedings of the IEEE/CVF Conference on Computer Vision
  and Pattern Recognition Workshops}, pp.\  1326--1334, 2019.

\bibitem[Codevilla et~al.(2018)Codevilla, Müller, López, Koltun, and
  Dosovitskiy]{codevilla2018endtoend}
Codevilla, F., Müller, M., López, A., Koltun, V., and Dosovitskiy, A.
\newblock End-to-end driving via conditional imitation learning, 2018.

\bibitem[Cunningham et~al.(2015)Cunningham, Galceran, Eustice, and
  Olson]{7139412}
Cunningham, A.~G., Galceran, E., Eustice, R.~M., and Olson, E.
\newblock Mpdm: Multipolicy decision-making in dynamic, uncertain environments
  for autonomous driving.
\newblock In \emph{2015 IEEE International Conference on Robotics and
  Automation (ICRA)}, pp.\  1670--1677, 2015.

\bibitem[{Dosovitskiy} et~al.(2017){Dosovitskiy}, {Ros}, {Codevilla}, {López},
  and {Koltun}]{dosovitskiy2017carla}
{Dosovitskiy}, A., {Ros}, G., {Codevilla}, F., {López}, A.~M., and {Koltun},
  V.
\newblock Carla: An open urban driving simulator.
\newblock \emph{Conference on Robot Learning}, pp.\  1--16, 2017.

\bibitem[Duan et~al.(2020)Duan, Li, Guan, Sun, and Cheng]{Duan_2020}
Duan, J., Li, S.~E., Guan, Y., Sun, Q., and Cheng, B.
\newblock Hierarchical reinforcement learning for self-driving decision-making
  without reliance on labelled driving data.
\newblock \emph{{IET} Intelligent Transport Systems}, 14\penalty0 (5):\penalty0
  297--305, feb 2020.

\bibitem[Fan et~al.(2023)Fan, Ma, Dai, Tan, and Low]{fan2023fedhql}
Fan, F.~X., Ma, Y., Dai, Z., Tan, C., and Low, B. K.~H.
\newblock Fedhql: Federated heterogeneous q-learning.
\newblock In \emph{Proceedings of the 2023 International Conference on
  Autonomous Agents and Multiagent Systems}, AAMAS '23, pp.\  2810–2812,
  Richland, SC, 2023. International Foundation for Autonomous Agents and
  Multiagent Systems.
\newblock ISBN 9781450394321.

\bibitem[Fan et~al.(2021)Fan, Ma, Dai, Jing, Tan, and Low]{fan2021fault}
Fan, X., Ma, Y., Dai, Z., Jing, W., Tan, C., and Low, B. K.~H.
\newblock Fault-tolerant federated reinforcement learning with theoretical
  guarantee.
\newblock \emph{Advances in Neural Information Processing Systems},
  34:\penalty0 1007--1021, 2021.

\bibitem[Gu et~al.(2016)Gu, Holly, Lillicrap, and Levine]{gu2016deep}
Gu, S., Holly, E., Lillicrap, T., and Levine, S.
\newblock Deep reinforcement learning for robotic manipulation with
  asynchronous off-policy updates, 2016.

\bibitem[Hoel et~al.(2020)Hoel, Driggs-Campbell, Wolff, Laine, and
  Kochenderfer]{8911507}
Hoel, C.-J., Driggs-Campbell, K., Wolff, K., Laine, L., and Kochenderfer, M.~J.
\newblock Combining planning and deep reinforcement learning in tactical
  decision making for autonomous driving.
\newblock \emph{IEEE Transactions on Intelligent Vehicles}, 5\penalty0
  (2):\penalty0 294--305, 2020.

\bibitem[Isele et~al.(2018)Isele, Rahimi, Cosgun, Subramanian, and
  Fujimura]{8461233}
Isele, D., Rahimi, R., Cosgun, A., Subramanian, K., and Fujimura, K.
\newblock Navigating occluded intersections with autonomous vehicles using deep
  reinforcement learning.
\newblock In \emph{2018 IEEE International Conference on Robotics and
  Automation (ICRA)}, pp.\  2034--2039, 2018.

\bibitem[Kendall et~al.(2019)Kendall, Hawke, Janz, Mazur, Reda, Allen, Lam,
  Bewley, and Shah]{8793742}
Kendall, A., Hawke, J., Janz, D., Mazur, P., Reda, D., Allen, J.-M., Lam,
  V.-D., Bewley, A., and Shah, A.
\newblock Learning to drive in a day.
\newblock In \emph{2019 International Conference on Robotics and Automation
  (ICRA)}, pp.\  8248--8254, 2019.

\bibitem[{Kiran} et~al.(2021){Kiran}, {Sobh}, {Talpaert}, {Mannion}, {Sallab},
  {Yogamani}, and {Pérez}]{kiran2021deep}
{Kiran}, B.~R., {Sobh}, I., {Talpaert}, V., {Mannion}, P., {Sallab}, A. A.~A.,
  {Yogamani}, S., and {Pérez}, P.
\newblock Deep reinforcement learning for autonomous driving: A survey.
\newblock \emph{IEEE Transactions on Intelligent Transportation Systems}, pp.\
  1--18, 2021.

\bibitem[{Levine} et~al.(2016){Levine}, {Finn}, {Darrell}, and
  {Abbeel}]{levine2016end}
{Levine}, S., {Finn}, C., {Darrell}, T., and {Abbeel}, P.
\newblock End-to-end training of deep visuomotor policies.
\newblock \emph{Journal of Machine Learning Research}, 17\penalty0
  (1):\penalty0 1334--1373, 2016.

\bibitem[Levy et~al.(2019)Levy, Konidaris, Platt, and Saenko]{levy2019learning}
Levy, A., Konidaris, G., Platt, R., and Saenko, K.
\newblock Learning multi-level hierarchies with hindsight, 2019.

\bibitem[Lillicrap et~al.(2019)Lillicrap, Hunt, Pritzel, Heess, Erez, Tassa,
  Silver, and Wierstra]{lillicrap2019continuous}
Lillicrap, T.~P., Hunt, J.~J., Pritzel, A., Heess, N., Erez, T., Tassa, Y.,
  Silver, D., and Wierstra, D.
\newblock Continuous control with deep reinforcement learning, 2019.

\bibitem[Lu et~al.(2020)Lu, Xu, Zhang, Qian, and Zhou]{9241055}
Lu, Y., Xu, X., Zhang, X., Qian, L., and Zhou, X.
\newblock Hierarchical reinforcement learning for autonomous decision making
  and motion planning of intelligent vehicles.
\newblock \emph{IEEE Access}, 8:\penalty0 209776--209789, 2020.

\bibitem[Mnih et~al.(2015)Mnih, Kavukcuoglu, Silver, Rusu, Veness, Bellemare,
  Graves, Riedmiller, Fidjeland, Ostrovski, et~al.]{mnih2015human}
Mnih, V., Kavukcuoglu, K., Silver, D., Rusu, A.~A., Veness, J., Bellemare,
  M.~G., Graves, A., Riedmiller, M., Fidjeland, A.~K., Ostrovski, G., et~al.
\newblock Human-level control through deep reinforcement learning.
\newblock \emph{nature}, 518\penalty0 (7540):\penalty0 529--533, 2015.

\bibitem[{Mnih} et~al.(2016){Mnih}, {Badia}, {Mirza}, {Graves}, {Harley},
  {Lillicrap}, {Silver}, and {Kavukcuoglu}]{mnih2016asynchronous}
{Mnih}, V., {Badia}, A.~P., {Mirza}, M., {Graves}, A., {Harley}, T.,
  {Lillicrap}, T.~P., {Silver}, D., and {Kavukcuoglu}, K.
\newblock Asynchronous methods for deep reinforcement learning.
\newblock In \emph{ICML'16 Proceedings of the 33rd International Conference on
  International Conference on Machine Learning - Volume 48}, pp.\  1928--1937,
  2016.

\bibitem[{Nachum} et~al.(2018){Nachum}, {Gu}, {Lee}, and
  {Levine}]{nachum2018data}
{Nachum}, O., {Gu}, S., {Lee}, H., and {Levine}, S.
\newblock Data-efficient hierarchical reinforcement learning.
\newblock In \emph{32nd Conference on Neural Information Processing Systems
  (NeurIPS 2018)}, volume~31, pp.\  3303--3313, 2018.

\bibitem[Nair et~al.(2018)Nair, McGrew, Andrychowicz, Zaremba, and
  Abbeel]{nair2018overcoming}
Nair, A., McGrew, B., Andrychowicz, M., Zaremba, W., and Abbeel, P.
\newblock Overcoming exploration in reinforcement learning with demonstrations,
  2018.

\bibitem[{Naveed} et~al.(2020){Naveed}, {Qiao}, and
  {Dolan}]{naveed2020trajectory}
{Naveed}, K.~B., {Qiao}, Z., and {Dolan}, J.~M.
\newblock Trajectory planning for autonomous vehicles using hierarchical
  reinforcement learning.
\newblock \emph{arXiv preprint arXiv:2011.04752}, 2020.

\bibitem[{Paden} et~al.(2016){Paden}, {Cap}, {Yong}, {Yershov}, and
  {Frazzoli}]{paden2016a}
{Paden}, B., {Cap}, M., {Yong}, S.~Z., {Yershov}, D., and {Frazzoli}, E.
\newblock A survey of motion planning and control techniques for self-driving
  urban vehicles.
\newblock \emph{IEEE Transactions on Intelligent Vehicles}, 1\penalty0
  (1):\penalty0 33--55, 2016.

\bibitem[{Qiao} et~al.(2020{\natexlab{a}}){Qiao}, {Schneider}, and
  {Dolan}]{qiao2020behavior}
{Qiao}, Z., {Schneider}, J., and {Dolan}, J.~M.
\newblock Behavior planning at urban intersections through hierarchical
  reinforcement learning.
\newblock \emph{arXiv preprint arXiv:2011.04697}, 2020{\natexlab{a}}.

\bibitem[{Qiao} et~al.(2020{\natexlab{b}}){Qiao}, {Tyree}, {Mudalige},
  {Schneider}, and {Dolan}]{qiao2020hierarchical}
{Qiao}, Z., {Tyree}, Z., {Mudalige}, P., {Schneider}, J., and {Dolan}, J.~M.
\newblock Hierarchical reinforcement learning method for autonomous vehicle
  behavior planning.
\newblock In \emph{2020 IEEE/RSJ International Conference on Intelligent Robots
  and Systems (IROS)}, pp.\  6084--6089, 2020{\natexlab{b}}.

\bibitem[{Schulman} et~al.(2017){Schulman}, {Wolski}, {Dhariwal}, {Radford},
  and {Klimov}]{schulman2017proximal}
{Schulman}, J., {Wolski}, F., {Dhariwal}, P., {Radford}, A., and {Klimov}, O.
\newblock Proximal policy optimization algorithms.
\newblock \emph{arXiv preprint arXiv:1707.06347}, 2017.

\bibitem[{Vinyals} et~al.(2019){Vinyals}, {Babuschkin}, {Czarnecki}, {Mathieu},
  {Dudzik}, {Chung}, {Choi}, {Powell}, {Ewalds}, {Georgiev}, {Oh}, {Horgan},
  {Kroiss}, {Danihelka}, {Huang}, {Sifre}, {Cai}, {Agapiou}, {Jaderberg},
  {Vezhnevets}, {Leblond}, {Pohlen}, {Dalibard}, {Budden}, {Sulsky}, {Molloy},
  {Paine}, Çaglar {Gülçehre}, {Wang}, {Pfaff}, {Wu}, {Ring}, {Yogatama},
  {Wünsch}, {McKinney}, {Smith}, {Schaul}, {Lillicrap}, {Kavukcuoglu},
  {Hassabis}, {Apps}, and {Silver}]{vinyals2019grandmaster}
{Vinyals}, O., {Babuschkin}, I., {Czarnecki}, W.~M., {Mathieu}, M., {Dudzik},
  A., {Chung}, J., {Choi}, D.~H., {Powell}, R., {Ewalds}, T., {Georgiev}, P.,
  {Oh}, J., {Horgan}, D., {Kroiss}, M., {Danihelka}, I., {Huang}, A., {Sifre},
  L., {Cai}, T., {Agapiou}, J.~P., {Jaderberg}, M., {Vezhnevets}, A.~S.,
  {Leblond}, R., {Pohlen}, T., {Dalibard}, V., {Budden}, D., {Sulsky}, Y.,
  {Molloy}, J., {Paine}, T.~L., Çaglar {Gülçehre}, {Wang}, Z., {Pfaff}, T.,
  {Wu}, Y., {Ring}, R., {Yogatama}, D., {Wünsch}, D., {McKinney}, K., {Smith},
  O., {Schaul}, T., {Lillicrap}, T.~P., {Kavukcuoglu}, K., {Hassabis}, D.,
  {Apps}, C., and {Silver}, D.
\newblock Grandmaster level in starcraft ii using multi-agent reinforcement
  learning.
\newblock \emph{Nature}, 575\penalty0 (7782):\penalty0 350--354, 2019.

\bibitem[Watkins(1989)]{qlearning}
Watkins, C.
\newblock Learning from delayed rewards.
\newblock 01 1989.

\bibitem[Wei et~al.(2014)Wei, Snider, Gu, Dolan, and Litkouhi]{6856582}
Wei, J., Snider, J.~M., Gu, T., Dolan, J.~M., and Litkouhi, B.
\newblock A behavioral planning framework for autonomous driving.
\newblock In \emph{2014 IEEE Intelligent Vehicles Symposium Proceedings}, pp.\
  458--464, 2014.

\bibitem[{Zhu} \& {Zhao}(2021){Zhu} and {Zhao}]{zhu2021a}
{Zhu}, Z. and {Zhao}, H.
\newblock A survey of deep rl and il for autonomous driving policy learning.
\newblock \emph{arXiv preprint arXiv:2101.01993}, 2021.

\end{thebibliography}
\bibliographystyle{icml2023}


\end{document}